# Hierarchical Bi-level Multi-Objective Evolution of Single- and Multi-layer Echo State Network Autoencoders for Data Representations

Naima Chouikhi[a*], Boudour Ammar[a] and Adel M. Alimi[a]

[a]REGIM-Lab: REsearch Groups in Intelligent Machines,University of Sfax, National Engineering School of Sfax (ENIS), BP 1173, Sfax, 3038, Tunisia.

[*] Corresponding author: naima.chouikhi@enis.usf.tn

**Abstract**

Echo State Network (ESN) presents a distinguished kind of recurrent neural networks. It is built upon a sparse, random and large hidden infrastructure called reservoir. ESNs have succeeded in dealing with several non-linear problems such as prediction, classification, etc. Thanks to its rich dynamics, ESN is used as an Autoencoder (AE) to extract features from original data representations. ESN is not only used with its basic single layer form but also with the recently proposed Multi-Layer (ML) architecture. The well setting of ESN (basic and ML) architectures and training parameters is a crucial and hard labor task. Generally, a number of parameters (hidden neurons, sparsity rates, input scaling) is manually altered to achieve minimum learning error. However, this randomly hand crafted task, on one hand, may not guarantee best training results and on the other hand, it can raise the network's complexity. In this paper, a hierarchical bi-level evolutionary optimization is proposed to deal with these issues. The first level includes a multi-objective architecture optimization providing maximum learning accuracy while sustaining the complexity at a reduced standard. Multi-objective Particle Swarm Optimization (MOPSO) has been achieving noticeable makings in solving problems having conflicting objectives. MOPSO is used to optimize ESN structure in a way to provide a trade-off between the network complexity decreasing and the accuracy increasing. A pareto-front of optimal solutions is generated by the end of the MOPSO process. These solutions present the set of candidates that succeeded in providing a compromise between different objectives (learning error and network complexity). At the second level, each of the solutions already found undergo a mono-objective weights optimization to enhance the obtained pareto-front. The evolved ESN recurrent AEs (basic and ML) show a noticeable strength in providing more expressive data from original ones. They are applied for classification purpose. Empirical results ensure the effectiveness of the conceived AEs for noisy and noise free data.

## 1. Introduction

Machine learning includes numerous fields [24] [55] [13] [2]. Recurrent Neural Networks (RNNs) [3] [4] are among the well-known machine learning tools. They have been efficiently used for addressing complexities in a number of tasks. Generally those tasks are nonlinear, dynamic and time-varying [31] [61] [15]. Thanks to their ability of information processing, RNNs have been widely involved for complex operations. Gradient based-methods such as Back Propagation Through Time [36] [10] [35] are the training algorithms the most used by traditional RNNs [44] [51].

Although they have realized successes in dealing with several problems, they are likely exposed to vanishing or exploding gradient problems. In fact, the error signal, once obtained and retro-propagated through layers, may disappear or increase exponentially. Consequently, the learning risks to diverge and to give unsatisfactory results.

Echo State Network (ESN) is a distinguished variant of RNNs [33]. It is characterized by a capacious untrained recurrent hidden space. It is called Dynamic Reservoir (DR). It typifies scattered synaptic links and linear readouts [27]. ESN is known to be one of the most effective recurrent models whenever there is a complex dynamic problem or a chaotic forecasting task to be handled. Within an ESN, an elaborate DR is created to gather various features extracted from the inputs. Once flown throughout the reservoir, these features are passed towards the readout map to generate the outputs. ESNs have fulfilled prosperous results when applied to several applications such as time series predictions [17], classification [33] and speech processing [54], etc.

Regarding the potential proven by ESN, it is used to deal with feature extraction problems in this work. Sometimes, original data representations [9] may not provide the most expressive data for a well-determined task. In literature, AEs [60] are known to be among the most efficient methods used to extract features from original ones [58]. The idea behind AEs is that the inputs are equal to the outputs. Throughout the mission of copying inputs into outputs, the hidden layer(s) undergoes progressive non-linear transformations to create new data from the original ones. Generally, existing AEs are based on single layered or deep feedforward networks. In this paper, ESN is used as an Autoencoder (AE). In order to mimic AEs, not only Basic ESN but also Multi-Layer ESN (ML-ESN) are used as Recurrent Autoencoders (RAEs). ML-ESN is a deep ESN having progressively interconnected reservoirs not just one. It was recently proposed in [42].

The challenge faced by AEs generally and ESN-RAE (basic and ML) especially is usually related to their design which is mainly related to their architecture and weights distribution. Creating a compact, equal or extended new data features is related to both of the AE's complexity and accuracy. Sometimes, a compact representation may lead to lose some details about the original data. It can occur, that extended representation engenders on complexity raise. Many scenarios can take place.

Indeed, ESN (basic and ML) has some limits especially related to the setting of its architecture and some weights parameters [18] [16]. In fact, some reservoir(s) properties can be hardly understood. A number of manually set parameters need tailored tuning. A brute-force is paid in a way to maximize the efficiency of ESN models. A set of trials-errors are performed in order to define suitable architecture for a targeted task. Nevertheless, ESN's parameters such as the number of hidden neurons, the input and inner connectivity rates (the number of non-zero synaptic connections) are still determined stochastically beforehand. For instance, in ML-ESN, the number of hidden reservoirs, the neurons within each reservoir and the synaptic weights relating the reservoirs to each others are randomly set.

In view of this stochastic aspect, neither the establishment of a well-tailored model would be sufficiently accurate nor the inner dynamics of the network would be strong. Consequently providing an optimal reservoir(s) for a given problem is crucial. While optimizing the reservoir(s), the complexity of the network should be taken into consideration. Setting an optimal architecture while maintaining the complexity at a low level are the challenges to be led off throughout this work. Encouraged by the achievements of multi-objective optimization approaches in machine learning [40], this work employs a Pareto-based multi-objective algorithm [25] to evolve the connectivity as well as the size of the reservoir(s).

In fact, multi-objective problems (MOPs) are often tackled by assembling all the objectives within a function. Then, standard methods are applied to that function. It is referred to an indirect single-objective problem. In some ways these approaches can be efficient but what if the studied problem doesn't support the aggregation of the objectives. Seen that this kind of optimization may not fit the given problem, a privileged way is to really carry out a multi-objective optimization (MOO). It means to give to each objective its merit when undergoing the optimization process. As a solution, Pareto-based approaches render optimal trade-offs between two or more different, not obligatory overlapping aims. This trade-off is traduced by a set of optimal front or fronts representing optimal solutions for a specific problem [32]. These solutions furnish a compromise between all the objectives. Thereafter, the final solution is picked out from this ensemble. A solution is considered to be Pareto-optimal if it cannot be enhanced in any objective while non-getting worse in at least one other goal.

A plenty of real-world optimization tasks involve a number of objectives. The mathematical formulation of a typical multi-objective optimization problem (MOOP) can be squeezed out such in (1).

$$\textit{\textbf{minimize/maximize }} F(x) = (f_1(x), \ldots, f_m(x))^T; \text{ s.t } x \in \Omega \qquad (1)$$

where $\Omega$ denotes the decision space and $x \in \Omega$ denotes a decision vector. $F(x)$ consists of $m$ objective functions $f_i : \Omega \rightarrow \mathrm{R}$, $i = 1,\ldots,m$, where $\mathrm{R}^m$ is the objective space.

The objectives in (1) are often conflicting. Restricting the focus on improving one objective may threaten the role of another one. Hence, it is hard to find a single solution that can bring up good result in terms of all objectives simultaneously. Instead, the trade-off between all the objectives can provide a set of optimal solutions constituting what we call the Pareto optimal front. The Pareto concept was proposed by Worth and Pareto [32] [48], and is defined according to definitions 1 and 2.

**Definition 1.** *Let $u = (u_1,\ldots,u_m)^T$ and $v = (v_1,\ldots,v_m)^T$ two vectors of size m. u is said to dominate v, denoted as u < v if $\forall i \in \{1,\ldots,m\}$, $u_i \leq v_i$ and $u \neq v$.*

**Definition 2.** *Let $x^* \in \Omega$ denotes a possible solution of problem (1). $x^*$ is called a Pareto optimal solution, if it $\nexists\ y \in \Omega$ such that $F(y) < F(x)$. The ensemble of the Pareto optimal solutions is named Pareto set (PS). It is denoted in equation (2).*

$$PS = \{x \in \Omega \mid \nexists\, y \in \Omega,\, F(y) < F(x)\}. \tag{2}$$

*Let the projection of the PS in the objective scheme be called the Pareto front (PF). This last is defined according to equation (3).*

$$PF = \{F(x) \mid x \in PS\}. \tag{3}$$

Over the past few years, evolutionary multi-objective optimization (EMO) [19] has been a popular technique highly useful in application and research. It has demonstrated considerable ability in solving complicated multiple objective problems. This kind of algorithms has been successful in handling more than just one objective test problems. They are based on a population of solutions navigating in a specific search space. From a step to another, a new population of solutions evolves. The new population is defined in terms of overall objectives. Numerous are the reasons behind their popularity. Not only they are relatively flexible and simple to implement but also they have a wide-spread applicability. Multi-objective particle swarm optimization (MOPSO) [21] has realized successes in puzzling out solutions in complex space. It is known by its efficient results as well as its rapid convergence.

In the literature, the problem of neural network parameters optimization has been tackled in many works such as [12] [23] [5] [46] [49] [43] [28] etc. Bouaziz *et al.* [12] proposed a hierarchical mono-objective optimization of Beta-Basis neural network. Although they came up with results improvement, finding one "optimal" solution in terms of one objective may result on complexity increase. Ammar *et al.* [5] used a multi-agent system to deal with neural network optimization. This method is very complex. It rises both of time and memory consuming. In ESN, all the weights are non-trained except the output ones. The non-trained weights are determined randomly before the learning process. In order to boost their performances in terms of accuracy, an evolutionary optimization of ESN's untrained weights is performed on the already obtained solutions from the PF from the network architecture optimization [17].

The proposed approach consists of a hierarchical bi-level EMO is performed for ESN-RAE (basic and ML). A multi-objective architecture optimization followed by a mono-objective weights optimization are carried out. The already evolved ESN (basic and ML) is used as RAE. On one hand, the high non-linearity and the rich dynamics provided by ESN are exploited in extracting more efficient features. On the other hand, a multitude of solutions (ESN-RAEs) realizing a compromise between the precision and complexity of the RAE are provided through a multi-objective evolutionary optimization. The outline of the paper is as follows. In section II, an overview about ESN (basic and ML) and how it can be used as RAE is performed. Evolutionary single and MO optimization is presented in section III. Single and MOPSO are detailed as they are the tools used in this work. In section IV, the hierarchical bi-level evolutionary optimization of ESN-RAE is thoroughly described. In section V, the already proposed approach is tested on several classification benchmarks. It attests its effectiveness throughout the empirical study. After all, a conclusion and outlooks for coming work are given.

## 2. Basic and Multi-Layer Echo State Network (Basic and ML): ESN as Autoencoder

RNNs get highly popular owing to their rich dynamics merged with remarkable nonlinear properties. In spite of their success in dealing with several academic tasks, RNNs suffer from a number of limitations such as gradient descent training methods. For instance, even the update of one parameter in a RNN is greedy in terms of computation expenses. On the other side, ESN is a recent recurrent architecture based on a very simple training method. It is designed for nonlinear systems modeling [11].

### 2.1. Basic and ML Echo State Network (ESN)

Typically, an ESN is characterized by a wide sparsely connected hidden layer called reservoir, within which the inner weights are randomly set and remain unchanged thereafter. The training is performed only for readout weights known by weights connecting the hidden reservoir to the output layer. Only those connections are updated generally by linear regression approaches. The remaining weights are non-trained. Due to the abovementioned specifications, ESN attracts ceaseless attention in supervised applications [30].
ESNs were first proposed by Jaeger and Hass [33] in order to predict chaotic time series. A wider-than-normal hidden layer resulting on what we call dynamic reservoir is randomly designed. Weights whose destination resides in the reservoir are randomly generated then still fixed and non-adapted to the task at hand. Only, the output weights are trained.

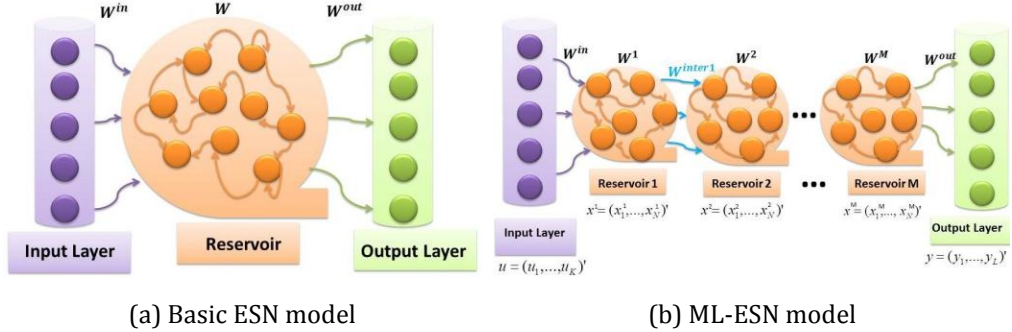

(a) Basic ESN model  (b) ML-ESN model

Figure 1: Basic and ML ESN architectures.

ESN is known only as a single layer recurrent network until Malik *et al.* [42] proposed Multi-Layer Echo State Machine in which a multitude of reservoirs are progressively interconnected. In this brief, both of Single and ML ESNs are studied. Their architectures are visualized in Figure 1.

The difference between the two architectures is at the level of the hidden layer, the first with one reservoir whereas in the second there is more than just one. The dynamics of ML-ESN are defined in equations (4)-(6).

$$x^1(n+1) = f(W^{in} u(n+1) + W^1 x^1(n)) \qquad (4)$$

$$x^k(n+1) = f(W^{inter(k-1)} x^k(n+1) + W^k x^k(n)) \qquad (5)$$

$$\vdots$$

$$x^M(n+1) = f(W^{inter(M-1)} x^M(n+1) + W^M x^M(n)) \qquad (6)$$

The output activation is computed according to equation (7).

$$y(n+1) = f^{out}(W^{out} x^M(n+1)) \qquad (7)$$

The mutual parameters between both of them are the input and output sequences denoted by *u* and *y*. $W^{in}$ and $W^{out}$ are respectively the input and output weight matrices. For single layer ESN (basic), *x* is the hidden state of the lonely reservoir whose recurrent inner weight matrix is denoted by $W^1$. Whereas, ML-ESN is distinguished by a set of $M$ heterogeneous reservoirs progressively interconnected via weights matrices $W^{inter(k)}$. This last relates the $k^{th}$ reservoir to the $(k+1)^{th}$ one, $k = 1..M-1$. To each hidden reservoir $R_i$ is associated an inner recurrent matrix $W^i$, $i = 1..M$. The reservoirs can be of equal or different sizes. $x^i$ designates the state vector of the $i^{th}$ reservoir $R_i$.

For the basic ESN with just one reservoir, just take M equal to 1. ESN (Basic and ML) has been applied for various tasks such as time series prediction, classification, pattern recognition, etc. It has realized considerable success in dealing with such kinds of problems. This strength and rich dynamics encouraged its use for feature extraction in this work. In fact, original data forms may be sometimes non capable to encompass all the specifications of the task to be resolved. Exploiting the original data representations to extract new more efficient features is highly recommended. The newly extracted features are very likely to give birth to novel hidden details which are more expressive than the traditional ones. In the literature, the most known approaches that perform in this type of tasks are Autoencoders (AEs) [60].

*2.2. Basic and ML Echo State Network Autoencoder*

Typical AE [60] is a neural network used for data coding according to unsupervised learning technique. To do so, AEs focus on learning the original data distribution in order to extract more efficient new features. The idea behind this kind of networks is that it attempts to copy its inputs to its outputs. It means making the outputs equal to the inputs. Typically, it is based on a smaller hidden layer size. But even when the hidden neurons number is large, it is still possible to discover good structures. It can be done by imposing some constraints such as sparsity. If a sparsity constraint is assessed at the level of hidden neurons, the AE would still getting efficient data structures.

AEs are mainly based on two main parts. The first phase is known as the encoding and the second is called the decoding. For Single-Layer AE, both of the encoder and the decoder are put together within a symmetrical framework (see Figure 2).

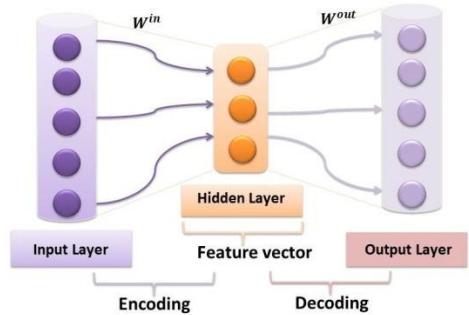

Figure 2: Autoencoder structure.

According to this figure, the synaptic connections relating the inputs to the hidden layer represent the encoder. The decoder is designated by the weights between the hidden units and the readout ones. In view of its achievements in dealing with several machine learning problems, ESN is used, throughout this paper, for creating new data representations. It plays the role of a recurrent AE.

Inspired by the concept of AE, an ESN-based AE is proposed in this section. The ESN preserves its normal structure. The outputs are set to be equal to the inputs.

Basically, ESN training is performed in two phases. The first step consists of a random feature mapping of the inputs. The second is a linear parameters solving. In fact, the hidden layer, once initialized, maps the input data into another feature space. This mapping is provided by the non-linear transformation when applying the hidden activation function (see Figure 3). ESN is distinguished by its random feature mapping as it uses, in general, the tanh function. Thus, it differs from other learning machine techniques which use usually kernel functions to create data features.

The input data, once processed throughout ESN-RAE (Basic and ML), the activation states of the $M^{th}$ reservoir are extracted and recorded in a state matrix a part. This last becomes the new representation of the already processed data. To be efficient, the new data should be well extracted thus the hidden states ought to be well computed.

To achieve this goal, ESN-RAE should be well designed at the level of architecture as well as training. As the Dynamic Reservoir (DR) is the fundamental construct of ESN, its design ought to be well crafted.

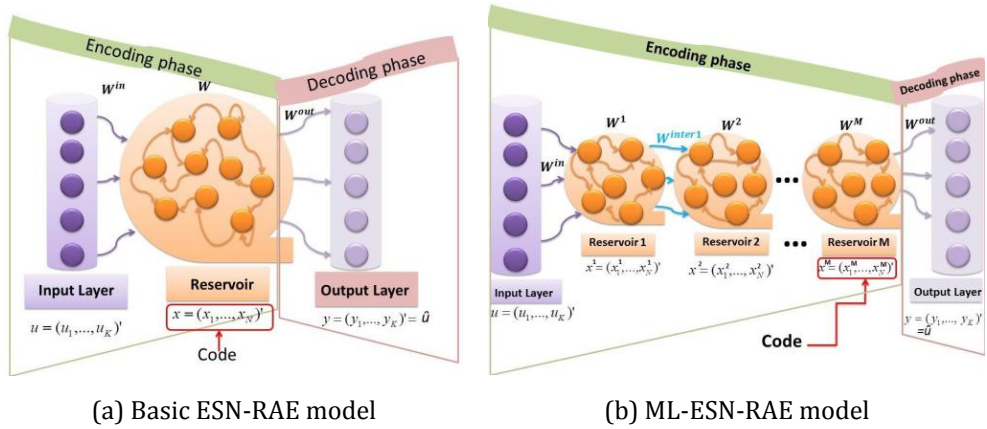

(a) Basic ESN-RAE model          (b) ML-ESN-RAE model

Figure 3: Basic and ML ESN-RAEs architectures.

Generally, both of the number of hidden units and the non-zero synaptic connection intervene on the DR's design. Let the connectivity rate designates the degree of non-zero connections regarding the total connections number. Typically, ESN design is non-constrained to precise formulas or rules [34]. There is just a recommendation to use a large reservoir within which the neurons are sparsely connected. Although a big DR may bring up high precision results, it can result on complexity raise. Therefore, providing optimal network parameters is obviously important not only to enhance the generalization capabilty but also to minimize the computation complexity. Commonly, the setting of the DR size seems to be among the most challenging tasks. In fact, the number of hidden neurons determines the connection weights' amount within the reservoir. Furthermore, it monitors the network's memory capacity.

The random setting of ESN (basic and ML) parameters is non-guaranteed. Even if it may bring up good accuracy results, there is no proof that the best architecture or weights are chosen. In fact, many are the objectives to be taken into consideration while setting the network. Maximizing the accuracy while minimizing the complexity are among these main objectives. Facing a huge random ESN setting, the solution proposed here is to use MO algorithms in order to provide more suitable network design well fitted to the targeted task. Seen the ceaseless success realized by EAs, they are chosen to be the optimization technique used in this work.

## 3. Single and Multi-Objective Particle Swarm Optimization

*3.1. Multi-Objective Evolutionary Optimization*

There has been an increasing appeal in using EAs to solve MOPs for years [47]. They are known as evolutionary multi-objective optimization algorithms (EMOAs) [19]. Indeed, evolution-based optimization techniques are dissimilar to classical methodologies in several ways. In fact, gradient information is not always used in a MO algorithm search process. Thus, evolutionary techniques are direct search procedures thus they are efficient in coping with a plenty of optimization tasks. As it has been already said, an evolutionary optimization method defines more than just one solution in a single iteration. These solutions are updated from iteration to another. Hence, EMOAs are unlike most classical optimization algorithms which update one solution at each iteration (a point approach), they are based on a number of solutions constituting the population concept. They deal with many objective tasks by mimicking the fundamental evolution principles on a number of candidates by means of certain evolutionary operators (fitness assignment, selection, crossover, mutation and elitism) [28].

In fact, evolving a population furnishes a parallel processing capability to the optimization algorithm which helps it to achieve quicker computation search. Moreover, the diversity created within the population supplies an EA with a high exploration capacity in the quest for optimal solutions. Thereby, MOPs may be solved easily.

In several MOPs, it is hard to get the optimal PF, thus, an estimate to it is quite welcome. To do so, MOEAs have realized ceaseless successes in dealing with complex MOPs. In general, the main difference between EMOAs resides on the fitness assignment method [19]. However, the majority of them are a part of a Pareto-based family involving Pareto dominance approach to elicit optimal or pre-optimal non-dominated solutions from an ensemble of evolving solutions. Figure 4 presents an example of a PF containing the PF of non-dominated solutions in a bi-objective space.

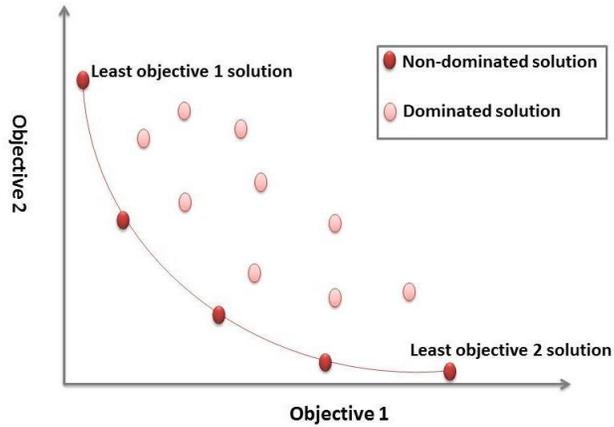

Figure 4: PF of non-dominated solutions in a bi-objective space.

Achieving a well-repartitioned and diversified PF is the primary goal of MOPs. There are many existing MOEAs [21] [63] [38] [22]. Indeed, they have been successfully exploited for solving MOPs. Particularly, MOPSO has been known by its capacity and effectiveness in dealing with complex optimization tasks. It has ensured more improved exploration and exploitation aspects provided by both local and global space search capability. Besides, the high speed as well as the quick convergence of single objective PSO attracted researchers to develop multi-objective optimization algorithms based on PSO [32]. More details about MOPSO are furnished in the coming section.

*3.2. Single and Multi-Objective Particle Swarm Optimization*

PSO [37] is among the well-known swarm intelligence (SI) techniques. It is derived from the conduct of either birds or fishes when displacing from a place to another. It is a population-based algorithm. An ensemble of particles flying in a precise search space constitutes what we call the swarm or the population. The particles' nature is strongly dependent on the problem to be studied [26].

There is a pair composed of position and velocity ($p_i$, $v_i$) assigned to every particle in the swarm. This pair is updated at each iteration according to the particle as well as the leader's behavior at the previous iteration. The leader is the particle having the minimum fitness value after the computation of its objective function. In fact, the swarm moves within a well-defined space. Each particle *i* is characterized by a velocity vector $v_i$. This last manages its movement within the space. At each epoch *it*, the evaluation of all the particles is performed according to a quality measure. This measure corresponds to the fitness or objective function [16].

In this kind of optimization, all the particles tend to either maximizing or minimizing their fitness function. Every particle records its local best position $lbp_i$. This latter corresponds to its best fitness achieved so far. As far as that goes, the global best position $gbp$ reached somewhere by a particle in the population is saved.

The velocity and the position vectors of all the swarm candidates are obtained, at each time step $it$, according to equations (8) and (9), respectively.

$$v_i(it+1) = wv_i(it) + C^1 rand^1_i (lbp_i(it) - p_i(it)) + C^2 rand^2_i (gbp(it) - p_i(it)) \quad (8)$$

$$p_i(it+1) = p_i(it) + v_i(it+1) \quad (9)$$

The inertia weight $w$ aims to create a balance between both exploration and exploitation processes. It manages the influence of the last time step velocity on the current time step one. Its value is generally in the interval [0,1]. The cognitive and social factors are denoted by $C^1$ and $C^2$, respectively. Their role consists of monitoring the particles' shifting towards the local or the global optima. It means that they supervise the trade-off between the space's exploration and exploitation. The matrices $rand^1_i$ and $rand^2_i$ contain positive values.

In order to be efficient in solving MOPs, the original PSO ought to be somehow altered. When settling a MOP, three key goals are to be fulfilled [18]. The first purpose is to rise up the number of pareto solutions. The other one aims to bring the obtained PF closer to the real one (admitting its location is known in advance). The last includes the enhancement of the solutions repartition throughout the search space.

The difference between PSO and MOPSO is that in this last, there is no longer a single leader for all the particles [20]. In fact, each particle chooses the corresponding leader as there will be several leaders. The PF is determined at the end of each iteration. The solutions provided by the PF are called leaders. They are saved in an external repository to be used in the coming time step. Optimal solutions found at time i$t$ are compared to those retrieved at i$t$ + 1. The comparison is mainly based on non-dominance criteria. It means if a solution, already within the PF, is dominated by another in the new PS, then it is discarded from the repository and replaced by the other one. The archive is limited by a pre-defined size. The pseudo-code of MOPSO is provided in algorithm 1.

| Algorithm 1: MOPSO algorithm |
|---|
| 1: Initialize the population |
| 2: Initialize the repository leaders |
| 3: Compute the objectives of all leaders |
| While (t < t-max) |
| For every particle in the swarm |
| 4: pick out a corresponding leader |
| 5: Compute the velocity (eq(4)) |
| 6: Compute the position (eq(5)) |
| 7: Apply a mutation operator if needed |
| 8: Update the particle's local best |
| End for |
| 9: Update the repository leaders (non-dominated solutions) |
| 10: Calculate the quality of the leaders |
| End while |
| 10: Visualize the leaders of the external repository |

As shows Algorithm 1, the new treatments added by MOPSO are the external repository, the multitude of leaders and the mutation operator. This operator boosts the exploration aspect of this algorithm. Throughout this paper, we conceive a 2-level hierarchical design based on a MOPSO of structure parameters followed by a PSO for non-trained weights values.

4. **Hierarchical Multi-Objective Particle Swarm Optimization of Basic and ML-ESN-RAE**

Generally, ESN's structural parameters are determined manually based on a series of experiences and targeted task requirements. This aleatory setting may come up with good generalization results but who can prove that they are the best regarding many constraints.
Usually, random initializations are non-guaranteed. Squirting optimization techniques within the network helps considerably in achieving optimal or pre-optimal solutions with high accuracies. Seen that Neural Networks are constrained to many aspects, restricting the focus on just improving the precision of the network can result on complexity increase.

In fact, there are many objectives to be considered when designing a neural network generally and ESN especially. Minimizing error rates while keeping the complexity at a low level are the main goals to be provided when conceiving an ESN. As the complexity of ESN is mainly and directly related to the design of the reservoir(s), the determination of the best reservoir's topology constitutes the main concern of this work. In order to facilitate the task, Basic ESN-RAE can be considered as a special case of ML-ESN-AE with $M = 1$.

A bi-level hierarchical optimization is proposed. It is clearly visualized in Figure 5. After the initialization of a population of ML-ESN-RAEs with different structural parameters, the architecture design optimization is launched. Since this design is multi-constrained, a multi-objective optimization technique is used here which is MOPSO. It provides all the good compromises that can be made between the structural complexity of ML-ESN-RAE and its modelling performance.

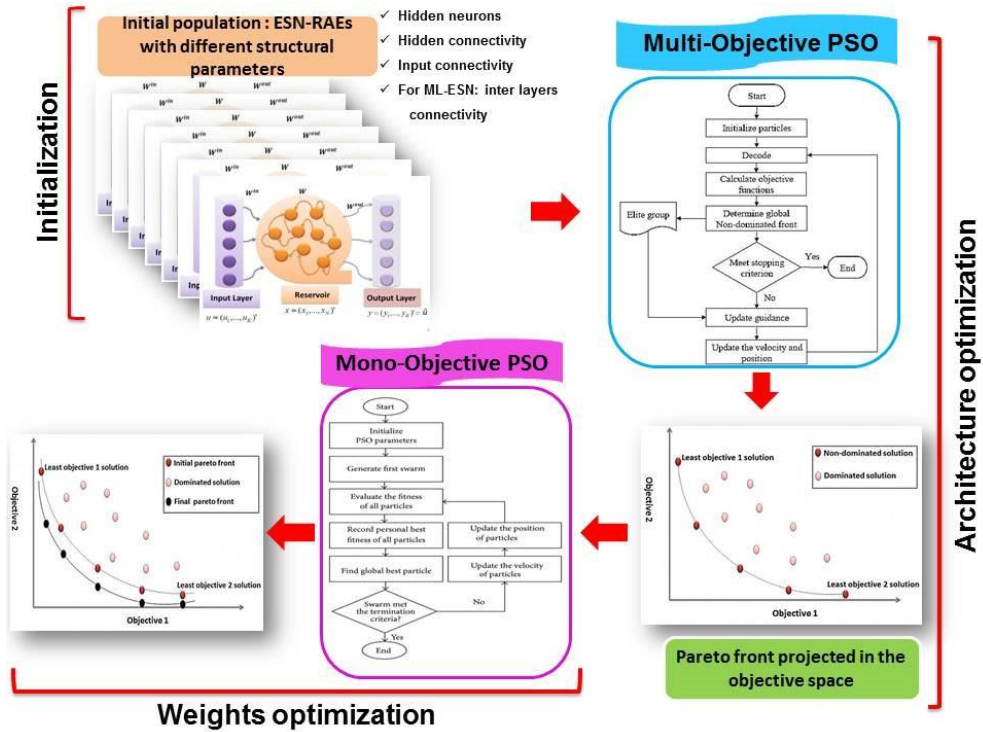

Figure 5: Hierarchical bi-level multi-objective evolutionary optimization of ESN-RAE.

*4.1. First level: Architecture optimization*

The reservoir(s) topology is traduced by its size as well as its connectivity rate. The network structure is encoded into multi-dimensional vector of floating-point numbers. The structural variables are the input, reservoirs' and inter-reservoirs' connectivity rates. The other terms are related to the reservoirs' sizes. Two kinds of MO are carried out. They are mainly based on the number of objectives to be studied. They are bi-objective and tri-objective optimizations.

- *Bi-objective PSO-ESN-RAE*

In the bi-objective optimization, the fitness function encompasses two sub-functions: the training RMSE and the Average Reservoirs Connectivity Rate (ARCR). Here the network's complexity is modeled by the ARCR. The reformulation of this kind of optimization is as follows:

*Particle* = {Reservoirs connectivity rates, Inter-Reservoirs connectivity rates, Input connectivity rate, Reservoirs sizes}
*Minimize F(Particle)={*Training RMSE, ARCR}
Subject to:

$1 \leq k^{th}$ reservoir size $\leq Rs^k$

$Lb-res^k < k^{th}$ reservoir connectivity rate $\leq Ub-res^k$

$Lb-in <$ Input connectivity rate $\leq Ub-in$

$Lb-inter^k <$ Inter-Reservoirs connectivity rate $\leq Ub-inter^k$

where $Rs^k$ denotes the maximum size of the $k^{th}$ reservoir. $Lb-res^k$ and $Ub-res^k$ are the lower and upper bounds of the $k^{th}$ reservoir connectivity rate, and $Lb-inter^k$ and $Ub-inter^k$ are lower and upper bounds of the inter weight matrix between the $k^{th}$ reservoir and the $(k + 1)^{th}$ one. $Lb-in$ and $Ub-in$ denote the lower and upper bounds of the input connectivity rate. The parameters already defined vary according to the task to be performed.

- *Tri-objective PSO-ESN-RAE*

In this optimization scenario, the bi-objective optimization, already described, is retaken with the addition of another objective which is the Average reservoir sizes (ARS). In fact, the complexity of the network is designated by two parameters to be minimized which are the ARS and the

ARCR. The only difference between both of scenarios is at the level of the fitness function which becomes:

*Minimize F(Particle)*={Training RMSE, ARCR, ARS} where the ARCR, the ARS and the training RMSE are computed according to equations (10), (11) and (12).

$$ARCR = \frac{\sum_{i=1}^{M} Reservoir\ Size(i)}{M} \quad (10)$$

$$ARS = \frac{\sum_{i=1}^{M} Reservoir\ connectivity\ rate\ (i)}{M} \quad (11)$$

$$RMSE = \sqrt{\frac{1}{p}\sum_{i=1}^{p}(u_i - \hat{u}_i)^2} \quad (12)$$

Where *M* is the number of layers within ML-ESN-RAE and *p* is the number of training patterns. $\hat{u}$ is the output of the ML-ESN-RAE corresponding to the input sequence *u*. This phase is crowned by obtaining a set of PF solutions as indicated in figure 5 to which the population converges.

*4.2. Second level: Weights optimization*

For further improvement of the PF, another mono-objective evolutionary weights optimization is applied to each solution within the PF. In such a case, one objective is favored compared to others. In this work, the precision designated by the training RMSE is selected as the fitness function to be minimized during weights optimization [18].

In fact, in ESN either basic or ML, the training is done only for the readout weights. The rest of the weights is non-trained while it is randomly initialized. Our idea consists on improving the accuracy of the obtained solution by optimizing the random non-trained weights ($W^{in}$, $W^{inter(k)}$ and $W^i$, $k = 1..(M − 1)$ and $i = 1..M$).

*4.3. Bi-level hierarchical optimization of basic and ML-ESN-RAE applied to classification*

In order to check the efficiency of the evolved basic and ML-ESN-RAE, the new encoded data is squirted into a classifier. In fact, each solution found by the end of the hierarchical optimization gives birth to specific features extracted from the original ones. These features correspond to RAEs reservoir states obtained after squirting every pattern from the dataset as input to the RAEs. A pseudo-code of the whole approach is presented in Algorithm 2. Let's call it "MOPSO-ML-ESN-RAE-PSO".

| Algorithm 2: MOPSO-ML-ESN-RAE-PSO algorithm |
|---|
| 1: Initialize the swarm: ML-ESN-RAEs with different architectural parameters. |
| 2: Undergo an MOPSO for the population in the swarm. |
| 3: Visualize the pareto front of optimal solutions. |
| For each solution in the pareto front |
| 4: Initialize the swarm: the concerned ML-ESN-RAE with different weights values parameters. |
| 5: Undergo a PSO for weights optimization according to one objective (RMSE). |
| 6: Obtain the optimal solution (optimal weight combination). |
| End for |
| 7: Obtain and visualize the enhanced pareto front: evolved ML-ESN-RAEs. |
| 8: Extract the $M^{th}$ reservoir states of each solution in the pareto front. |
| 9: Squirt the new data representation into a classifier. |
| 10: Compute the classification accuracy. |

Once obtained by ESN-RAE (basic or ML), the new data (reservoir states) are injected into a classifier. The classification accuracy is computed. It is compared to that obtained when entering the original data directly to the classifier without passing from the ESN-RAE (basic or ML). The choice of the convenient ESN-RAE (basic or ML) refers to the priority of the pre-fixed objectives in accordance with the task to be handled.

## 5. Experimental results

Empirical results are performed on several well-known classification datasets. For each test, a set of decision variables and objectives are determined. For all the datasets, bi- and tri-objective optimizations are carried out as it is already mentioned previously. A comparison between the performances of several EAs is effectuated. The empirical results for each dataset are presented as follows: in the first level, the obtained PF and the population candidates are presented. Then, in the second level, the enhanced non dominated PF solutions get after weights optimization by several EAs are visualized. Finally, numerical results in terms of classification accuracy are given.

The setting of the problems parameters as well as those for MOPSO and PSO are included in Table 1 and Table 2. For the first level, MOPSO is chosen thanks to its distinguished exploration and exploitation power. For the second level, a set of EAs are implemented for weights optimization. The implemented EAs include Differential Evolution (DE) [23], Firefly Algorithm (FA) [6], Artificial Bee Colony (ABC) [16], Harmony Search (HS) [29] and Biogeography-Based Optimization (BBO) [53]. The proposed approach is executed five times and the average precision is taken as the final result.

Table 1: Parameters of MOPSO and PSO.

| Method | Parameters | Value |
|---|---|---|
| MOPSO | $C_1$ | 0.1 |
| | $C_2$ | 0.2 |
| | Constriction factor $w$ | 0.5 |
| | Population size | 20 |
| | Number of leaders | 10 |
| | Mutation rate | 0.5 |
| | Number of iterations | 50 |
| PSO | $C_1$ | 0.1 |
| | $C_2$ | 0.2 |
| | Constriction factor $w$ | 0.9 |
| | Population size | 10 |
| | Number of iterations | 50 |

The setting of the MO evolved ML-ESN-RAE figures in Table 2. Once the final non-dominated solutions are retrieved, the hidden states of each solution in terms of RMSE are taken as the new data code. This new code is squirted into an SVM classifier [1] in order to make a classification. The Classification Accuracy (CA) of each solution is computed according to equation (13).

$$CA = \frac{Number\ of\ well\ classified\ patterns}{Total\ number\ of\ testing\ patterns} \quad (13)$$

The best solution is the one having the higher CA. The performances of ESN-RAE (basic and ML) are compared to those of the evolved ESN-RAE. The solution that gives the best accuracy is the winner as it gives birth to more efficient data representation.

Table 2: Multi-objective evolved ML-ESN-RAE initialization.

| Parameter | Value |
|---|---|
| M | 2 |
| $Lb-res^k$ | 0 |
| $Ub-res^k$ | 1 |
| $Lb-inter^k$ | 0 |
| $Ub-inter^k$ | 1 |
| $Lb-in$ | 0 |
| $Ub-in$ | 1 |

By the end of the algorithm, there are a number of optimized solutions. The choice is dependent on the priority of the objectives. In this work, precision is more prior. For other tasks, another objective may be the most prior. It mainly depends on the requirements of the task to be handled.

*5.1. Accuracy-based results*

*5.1.1. ECG200 dataset*

ECG200 dataset was proposed by Olszewski during his thesis [14]. It includes records of 200 heart beats signals. Every signal consists of 96 inputs. Each series traces the electrical activity recorded during one heartbeat. Two classes are defined here: normal and abnormal heartbeat. Figure 6 shows an example visualizing the obtained PF as well as the population after undergoing a bi-objective PSO of structural parameters of ML-ESN-RAE. The same work is done for basic ESN-RAE.

The solutions are partitioned throughout the objective space towards which a number of candidates are converging. Once the approach's first level designated by the bi-objective PSO of architectural parameters is achieved, the second level is launched. During this last, a mono-objective weights optimization is performed to enhance the retrieved PF solutions in terms of one objective (RMSE). Figure 7 shows the enhanced non-dominated solutions in terms of RMSE by a number of EAs where PSO realizes remarkable error minimization compared to the initial PF as well as to those given by other EAs. The same work is done for the tri-objective PSO.

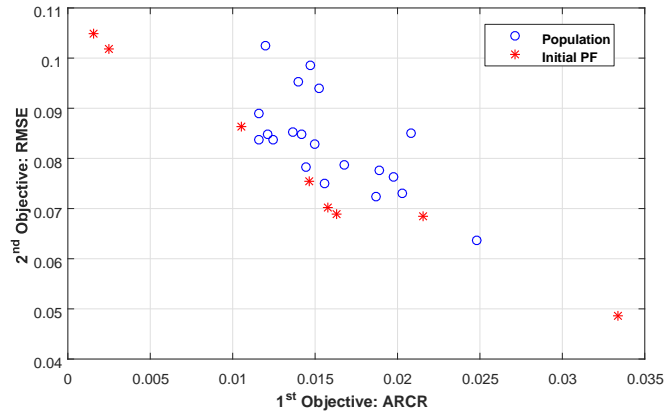

Figure 6: 1*st* level: Population (blue circles) convergence towards the obtained bi-objective PF (red stars) of ML-ESN-RAE.

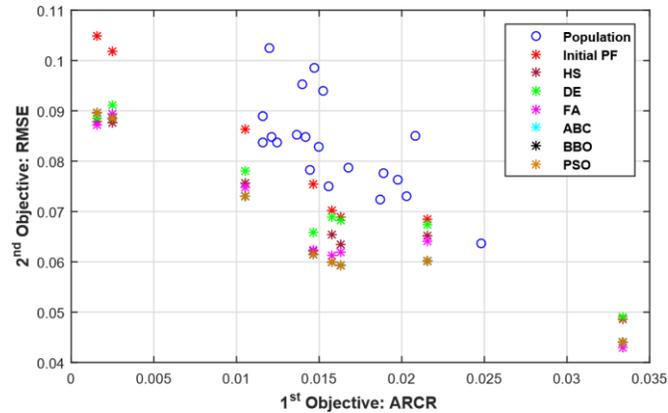

Figure 7: 2*nd* level: weights optimization of the obtained PF solutions by several EAs.

Figure 8 represents the population (tri-objective optimization), the initial PF (obtained at the 1*st* level) and the enhanced PF solutions in terms of RMSE by several EAs. The tri-objective PSO gives birth to a 3D PF in terms of the three objectives already enumerated in the previous section. Once the evolution process is finished and the final PF is obtained. The hidden activation states of the already selected solutions are squirted into the classifier as the new data representation. The classification accuracy is computed and the results are gathered in Table 3. The solutions' diversity permits to choose the equivalent one. In this study, the one having the maximum accuracy is picked out. Table 3 reports a comparison based on the testing classification accuracy of the studied approaches as well as other existent methods in the literature. Based on what reports Table 3, after undergoing an evolutionary optimization process, the performance of basic as well as ML-ESN-RAE are enhanced. They realized advanced accuracy results compared to other existent approaches. It is also clear from the table that the use of ESN-RAE to create new data code give better results than directly squirting the original data into SVM classifier.

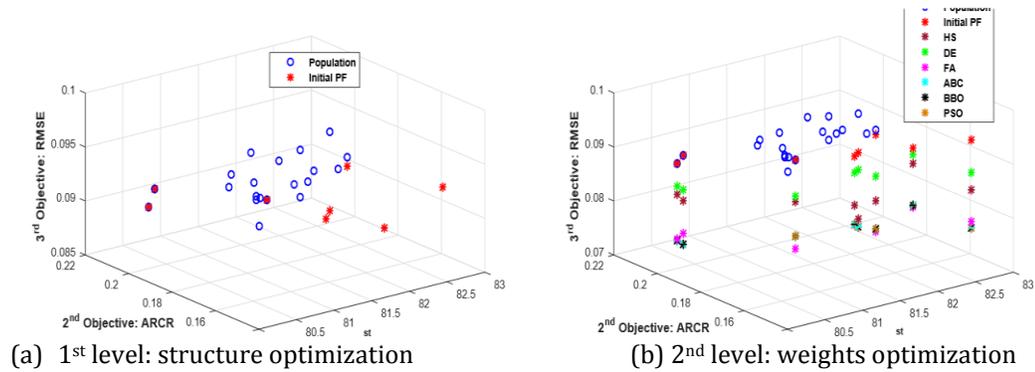

(a) 1$^{st}$ level: structure optimization    (b) 2$^{nd}$ level: weights optimization

Figure 8: 1$^{st}$ level: Population(blue circles)convergence towards the obtained bi-objective PF (red stars) of ML-ESN-RAE (left). 2$^{nd}$ level: weights optimization of the obtained PF solutions by several EAs (right).

Table 2: Classification accuracy-based comparison with other existent approaches on the ECG200 dataset.

| Method | CA |
|---|---|
| STMF [52] | 0.700 |
| SVM [62] | 0.790 |
| Swale [41] | 0.830 |
| SpADe [41] | 0.744 |
| GeTeM [41] | 0.800 |
| 1-NN [62] | 0.880 |
| CNN [62] | 0.980 |
| Jeong's method [62] | 0.840 |
| Gorecki's method [62] | 0.830 |
| LS [59] | 0.870 |
| FS [59] | 0.766 |
| ST+FCBF [59] | 0.766 |
| LPP [45] | 0.710 |
| NCC [52] | 0.770 |
| N5S2 [56] | 0.770 |
| EDTW [45] | 0.825 |
| N8S5 [56] | 0.840 |
| DSVM [45] | 0.855 |
| SVM | 0.816 |
| ESN-RAE | 0.846 |
| ML-ESN-RAE | 0.877 |
| *MOPSO-ESN-RAE-PSO (bi-objective)* | **0.895** ±0.022 |
| *MOPSO-ESN-RAE-PSO (tri-objective)* | **0.888** ±0.044 |
| *MOPSO-ML-ESN-RAE-PSO (bi-objective)* | **0.905** ±0.012 |
| *MOPSO-ML-ESN-RAE-PSO (tri-objective)* | **0.900** ±0.014 |

*5.1.2. Coffee dataset*

Two coffee bean varieties have attracted widely economic interest. They are Coffea Arabica and Coffea Canephora named also Robusta [8]. Commonly, Arabica species are more valued by the trade as they are characterized by a finer flavor. Both of coffee beans species can be distinguished regarding their sizes. However this visual criteria vanishes when processing coffee. Issues focused on discriminating the two species according to a chemical analysis. The dataset can be retrieved in the UCR Archive. This data base contains 286 inputs and two output classes.

Figure 9 represents the elicited PF as well as the population after going through both of bi- and tri-objective optimizations of ML-ESN-RAE structural parameter. Figure 10 reports the results brought after level 2. The enhancement of the obtained solutions after weights optimization is obviously clear.

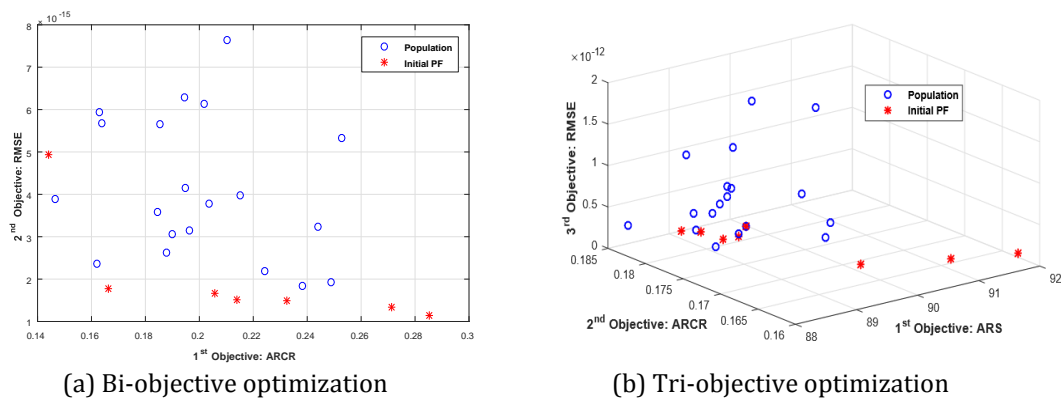

(a) Bi-objective optimization  (b) Tri-objective optimization

Figure 9: $1^{st}$ level: Population (blue circles) convergence towards the obtained bi-objective PF (red stars) of ML-ESN-RAE.

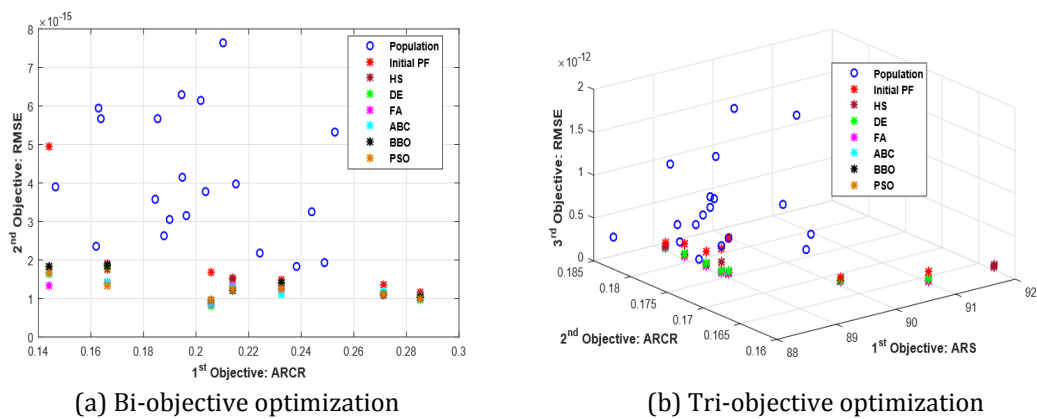

(a) Bi-objective optimization  (b) Tri-objective optimization

Figure 10: $2^{nd}$ level: weights optimization of the obtained PF solutions by several EAs.

The non-dominated solutions are distributed throughout the objective space constituting the initial PF. Candidates within the population are converging to the PF.

Table 4 highlights the efficiency of using evolutionary ESN-RAE and especially the impact of adding another hidden layer within the ESN for the three datasets. This is thanks to the high non-linearity provided by ESN-RAE's reservoir(s). Also, the effect of adding PSO either mono- or multi-objective is highly remarkable throughout the classification accuracy rise.

Table 4: Classification accuracy-based comparison with other existent approaches on the Coffee dataset.

| Method | CA |
|---|---|
| C.45 [8] | 0.822 |
| NN [8] | 0.858 |
| NB [8] | 0.858 |
| NNDTW [8] | 0.929 |
| Naive Bayes [41] | 0.929 |
| Swale [41] | 0.730 |
| SpADe [41] | 0.815 |
| GeTeM [41] | 0.857 |
| SVM | 0.894 |
| ESN-RAE | 0.919 |
| ML-ESN-RAE | 0.930 |
| ***MOPSO-ESN-RAE-PSO (bi-objective)*** | **0.971** ±0.014 |
| ***MOPSO-ESN-RAE-PSO (tri-objective)*** | **0.965** ±0.021 |
| ***MOPSO-ML-ESN-RAE-PSO (bi-objective)*** | **0.984** ±0.010 |
| ***MOPSO-ML-ESN-RAE-PSO (tri-objective)*** | **0.976** ±0.020 |

*5.1.3. Breast cancer dataset*

This benchmark was generated by Dr. Wolberg [7] based on consecutive patients. Data items are extracted from digitized images. They consist of 699 patterns where each pattern includes 9 features. They contain illustrations of breast mass Fine Needle Aspirate (FNA). The diagnosis result will be either malignant or benign. RS is equal to 500.

Figure 11 draws a plot of ML-ESN-RAEs population convergence towards the initial PF (obtained after MOPSO structure optimization) which is enhanced, in its turn, after weights optimization by PSO.

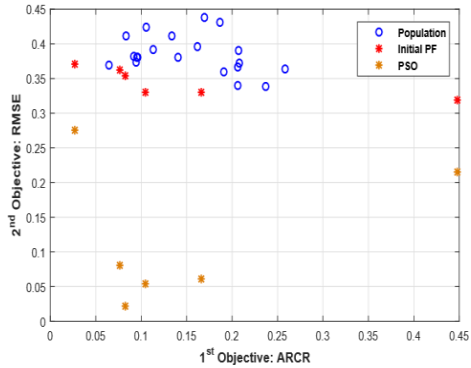 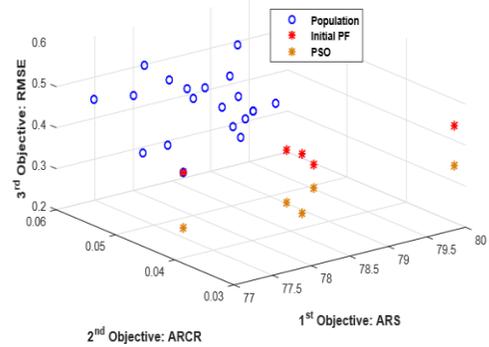

(a) Bi-objective optimization    (b) Tri-objective optimization

Figure 11: Population (blue circles) convergence towards the obtained bi- and tri-objective PF (red stars) of ML-ESN-RAE (left). PF enhancement throughout weights optimization by PSO (brown squares) (right).

Both of bi- and tri-objective optimizations are visualized. Table 5 reports the classification performance of the proposed RAEs before and after PSO evolution. It shows, as well, the accuracy achieved by other methods in the literature. According to table 5, the conceived approaches realized greater results than any of the other mentioned methods in terms of classification precision.

Table 5: Classification error-based comparison with other existent approaches on the Breast cancer dataset.

| Method | CA |
|---|---|
| C-RT [7] | 0.899 |
| NB [50] | 0.930 |
| CART with feature selection [39] | 0.946 |
| C.45 [7] | 0.948 |
| SVM | 0.920 |
| ESN-RAE | 0.935 |
| ML-ESN-RAE | 0.953 |
| **MOPSO-ESN-RAE-PSO (bi-objective)** | **0.983** ±0.002 |
| **MOPSO-ESN-RAE-PSO (tri-objective)** | **0.979** ±0.003 |
| **MOPSO-ML-ESN-RAE-PSO (bi-objective)** | **0.989** ±0.002 |
| **MOPSO-ML-ESN-RAE-PSO (tri-objective)** | **0.985** ±0.004 |

### 5.1.4. ECG Five Days dataset

Data is from a 67 years old male [14]. This dataset contains 23 patterns for the training and 861 patterns for the test. Each data pattern includes 136 inputs. Table 6 gives the CAs achieved by both of the designed RAEs and already designed methods in the literature.

Table 6: Classification error-based comparison with existent methods in the literature (ECG Five Days).

| Method | ECG Five Days |
|---|---|
| ED [56] | 0.797 |
| DTWR [56] | 0.797 |
| N5S2 [56] | 0.780 |
| 1NN ED [59] | 0.797 |
| 1NN DTW [59] | 0.768 |
| ST+CFS [59] | 0.971 |
| Swale [41] | 0.710 |
| SpADe [41] | 0.735 |
| 2DD [41] | 0.944 |
| EE [41] | 0.822 |
| ESN-RAE | 0.953 |
| ML-ESN-RAE | 0.972 |
| **MOPSO-ESN-RAE-PSO (bi-objective)** | **0.976** ±0.013 |
| **MOPSO-ESN-RAE-PSO (tri-objective)** | **0.968** ±0.026 |
| **MOPSO-ML-ESN-RAE-PSO (bi-objective)** | **0.974** ±0.014 |
| **MOPSO-ML-ESN-RAE-PSO (tri-objective)** | **0.971** ±0.024 |

Throughout the successive empirical results, many interpretations can be extracted. In fact, ESN ensures its efficiency as a recurrent AE either basic or ML. It is also revealed that the PSO-based evolution either mono- or multi-objective process has improved considerably the quality of the new extracted features thus the classification accuracy. ML-ESN-RAE is almost more stable than basic ESN-RAE. This last shows great results too. Further experiments are carried out. The classification accuracies per class are visualized.

Figure 12 represents examples of the confusion matrices of a number of datasets for both of basic and ML-ESN-RAEs. These matrices correspond to the best run among the five runs.

(a) ECG200: evolved ML-ESN-RAE.　　　(b) ECG200: evolved ESN-RAE.

(c) Coffee: evolved ML-ESN-RAE.　　　(d) Coffee: evolved ESN-RAE.

(e) Breast cancer: evolved ML-ESN-RAE.　　　(f) Breast cancer: evolved ESN-RAE.

Figure 12: Examples of Confusion matrices for one run of evolved basic and ML-ESN-RAEs in a number of datasets.

## 5.2. Comparison between EAs

The rest of the empirical tests include the bi-objective optimization case as it has given better results than the tri-objective one.

At the beginning of experiments, it was mentioned that PSO is compared to a set of existent meta-heuristics applied for weights optimization ($2^{nd}$ level). Throughout previous figures (7, 9, 11), it was shown that the implemented algorithms result on non-dominated solutions enhancement in terms of precision (RMSE). A numerical statistical study of the performances of those algorithms after squirting the new data within the classifier is carried out. Figure 13 presents the classification accuracy results on a number of datasets for both of evolved basic and ML-ESN-RAEs.

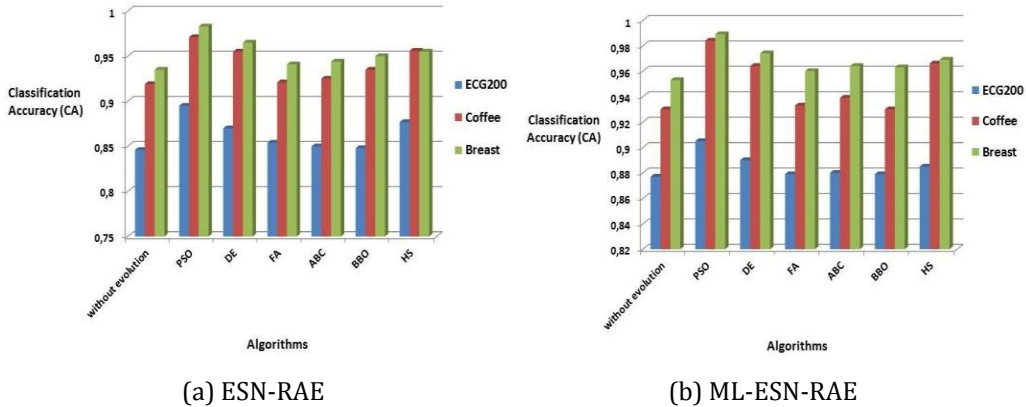

(a) ESN-RAE                    (b) ML-ESN-RAE

Figure 13: Classification accuracy-based comparison between several EAs on a number of datasets.

Figure 13 makes the fact of using PSO efficient and effective throughout the comparison included within it. In fact, the classification accuracy results realized by PSO weights optimization overtake those achieved by five other metaheuristic approaches of both basic and ML-ESN-RAE. ML-ESN-RAE brings up sturdier results than basic ESN-RAE. In order to boost the comparison and show further outperformance of PSO, the execution time is tracked for all of the algorithms. Then, each EA's execution time is divided by the sum of execution times of all of the algorithms.

Figure 14 gives statistical timing percentile results of each EA for a set of datasets in the case of ML-ESN-RAE. Another time, PSO outperforms the others as its execution time is faster than other studied methods.

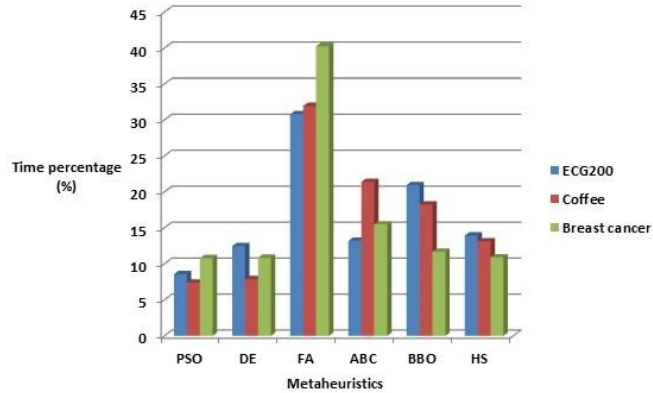

Figure 14: Timing percentage-based comparison between PSO and other metaheuristic algorithms.

### 5.3. Noise addition results

Real-world data, which constitute generally the input of Data Mining approaches, may be damaged by different breakdowns such as noise disturbing. Noise is considered as meaningless data disturbing any data mining analysis [57]. It is an unavoidable problem affecting the data aggregation process. Noise comes from either implicit measurement tools' errors or random faults generated by batch process.

The performance of any intelligent system in general and our systems in special heavily depends on the quality of the training data, but also on the robustness against noise. To deal with this issue, two noise levels are applied randomly to both of the training and testing databases. Let's define the SNR as the ratio between the original signal and the noise. It is expressed in dB. The two breakdown levels are picked out for SNR=50dB and SNR=10dB. The behavior of the studied approaches before and after including disturbances is tracked throughout statistical histograms. Figures 15 (a), (b) and (c) report the achieved results.

Based on what is visualized in figure 15, the bi-level hierarchical evolution has improved noticeably the robustness of the studied RAEs. The amelioration percentages after undergoing architecture and weights optimizations are included in Table 7.

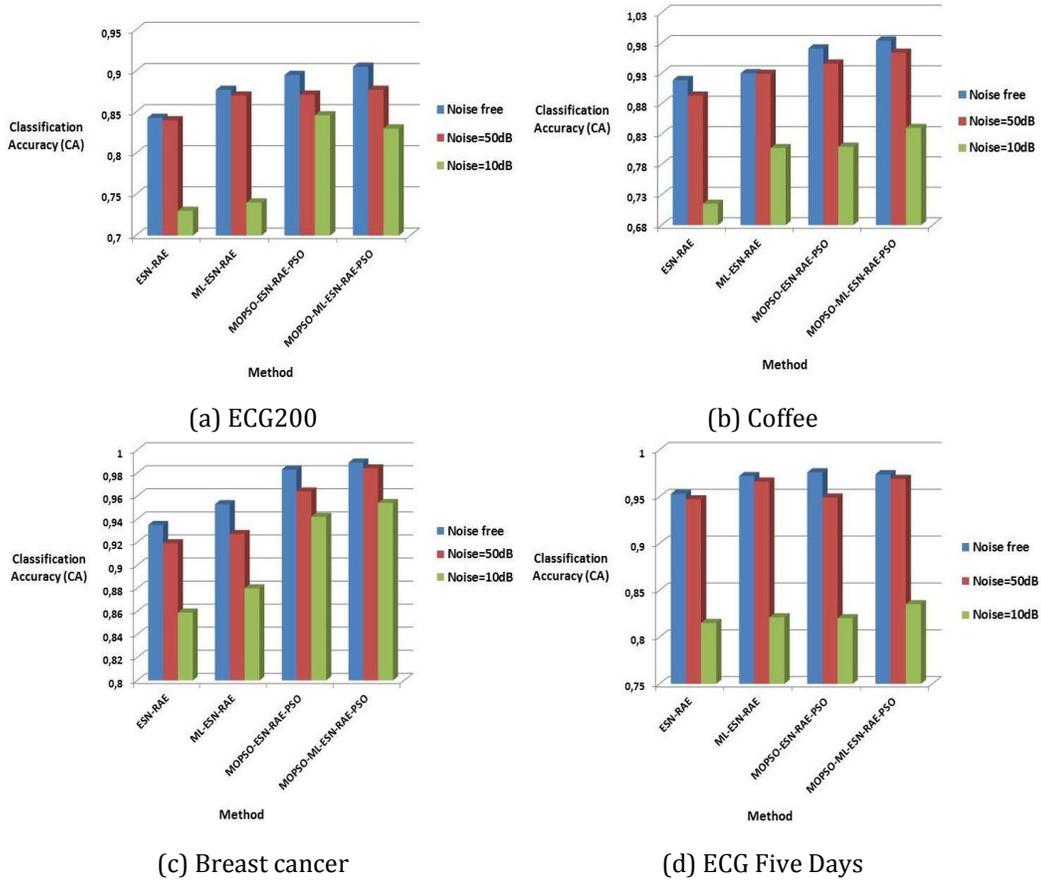

(a) ECG200  (b) Coffee

(c) Breast cancer  (d) ECG Five Days

Figure 15: Statistical analysis of the performance evolution of ESN-RAE, ML-ESN-RAE, MOPSO-ESN-RAE-PSO, MOPSO-ML-ESN-RAE-PSO with and without addition of Gaussian noise levels.

Table 7: Improvement percentages after undergoing architecture and weights optimization. ML-ESN-RAE (top) and basic ESN-RAE (down).

| Dataset | Noise=50dB | Noise=10dB |
|---|---|---|
| **ECG200** | 0.08 | 12.16 |
| | 3.60 | 15.9 |
| **Coffee** | 3.70 | 4.08 |
| | 5.93 | 13.14 |
| **Breast** | 6.14 | 8.40 |
| | 4.89 | 9.66 |
| **ECG Five Days** | 0.31 | 1.70 |
| | 0.21 | 0.61 |

In fact, the CA improvement rate after undergoing the hierarchical bi-level architecture and weights optimization is tracked for both of single and ML recurrent AEs.

In accordance with what brings up table 7, the robustness against noise is obvious after PSO-based evolution. Both of single and ML evolved ESN-RAEs have realized better classification accuracies even with break-downs addition.

### 5.4. Discussion

The choice of the datasets for experiments is not done randomly. We tried to choose different datasets at the level of dimension and dimensionality. The dimension is the number of data patterns whereas the dimensionality denotes the set of features that is used to situate a pattern in a predefined space. The approach is applied to databases having the dimension greater than the dimensionality (ECG 200, Breast cancer) and vice-versa (Coffee and ECG 200). For the training and testing data, we used datasets where the training patterns are less (ECG Five Days), equal (ECG 200 and Coffee) and greater (Breast cancer) than the testing data size. By means of the empirical results which are already presented, the conceived approach has a significant impact on the precision of the proposed recurrent AEs. The bi-level hierarchical evolutionary optimization has improved the performance results considerably in terms of precision, diversity and stability, etc.

#### 5.4.1. Multi-objectiveness and diversity

The most advantage of MOO is the fact that they come up with a number of solutions not just one as is the case of single objective optimization. The PF provides diverse solutions making a trade-off between a set of objectives. This diversity has been proven in that work through the already presented figures (6, 7, 8, 9, 10 and 11). By furnishing many solutions, the choice of the best solution is done according to the objectives priority of the studied task. In this work, the precision (RMSE) has been considered as the most prior objective. Thus, the weights optimization has enhanced more and more that objective. It is up to the task to be studied to choose the appropriate solution from a set on non-dominated solutions.

#### 5.4.2. Stability against noise

The purpose behind integrating noise within the training and testing datasets is to control the behavior of the system towards that disturbance.

Based on what reports section 5.3, the evolution process has ameliorated the power of basic and ML-ESN-RAEs. In fact, the CA realized after applying the bi-level hierarchical optimization is lightly affected compared to that obtained without optimization. Thus, the robustness enhancement of basic and ML-ESN-RAEs is a proof of the efficiency of the conceived approach in terms of stability.

*5.4.3. Single Layer vs Multi-Layer evolved ESN-RAEs*

Throughout the given experimental results, both of single and ML evolved ESN-RAEs have ensured high precision capabilities. ML architecture has surpassed the mono-layer one in terms of accuracy in the majority of the studied datasets. However, even the single layer model has sometimes overtaken the ML one such as with ECG Five Days dataset. Also, it has shown a distinguished power in the case of noise addition (Table 7 and Figure 14). The choice of what model fits more and brings better performance is mainly related to the task to be handled. Thus, either basic or ML, evolved ESN-RAE seems to be a very important paradigm providing efficient data representations.

## 6. Conclusion

In this paper, ESN is presented in both of single and multi-layer forms. The rich dynamics of ESN either single or ML are merged with the strengths of AEs in dealing with representation learning. This mixture results on ESN-RAE (basic and ML). Seen that the general purpose of AEs and especially this based on ESN is to provide more accurate and representative data from original ones, the design of the proposed recurrent AE is well studied. While tending to bring up sturdier and more precise results, the complexity of the studied intelligent model is taken into consideration. Seen that in some applications the complexity of the model goes side by side with precision, a compromise between those two objectives is provided throughout a multi-objective optimization. In fact, ESN (basic and ML) is known by considerable random parameters initialization. These parameters are mainly related to the architecture as well as the weights within this network. In order to furnish suitable ESN-RAE (basic and ML) parameters, at a first level, MOPSO is implemented to make a multi-objective optimization of the RAE architecture (reservoir(s) size, weights matrices connectivity rates, etc.). This step comes up with a set of non-dominated solutions constituting the PF. This last illustrates a trade-off between both of precision and complexity. At level 2, in order to enhance more and more the quality of the already obtained PF, each given solution undergoes weights optimization. At this level, the optimization is done according to one objective which is the accuracy. A single-objective PSO is elected to do this job. Once these steps are well achieved, the hidden activation states of the final pareto solutions (the new data representation) are squited to a SVM classifier. The classification accuracy is computed for each of the following scenarios:

- When original data are squirted directly to the classifier.

- When original data are pre-processed by the Basic and Multi-layer Echo State Network Autoencoder then the corresponding reservoir states are introduced to the SVM classifier.

- When original data are pre-processed by the evolutionary Basic and Multi-layer Echo State Network Autoencoder then the corresponding reservoir states are introduced to the SVM classifier.

Other metaheuristic algorithms are implemented at level 2 to be compared with PSO. The tests reveal the potential of PSO in terms of precision and time. Also, the obtained results are compared with other existent literature approaches for classification purposes. The proposed models show a distinguished advance compared to those methods. By the end, The proposed evolved RAEs show a stability and robustness improvement compared to the non-evolved ones. This enhancement is noticed throughout random Gaussian noise addition to the datasets. Other variants of evolutionary basic and ML ESN-RAE are under investigation to improve the results already obtained further.

**Acknowledgment**

The research leading to these results has received funding from the Ministry of Higher Education and Scientific Research of Tunisia under the grant agreement number LR11ES48.